\documentclass{article} % For LaTeX2e
\usepackage{iclr2027_conference,times}

\usepackage{amsmath,amsfonts,bm}

\def\eqref#1{equation~\ref{#1}}
\def\1{\bm{1}}

\DeclareMathAlphabet{\mathsfit}{\encodingdefault}{\sfdefault}{m}{sl}
\SetMathAlphabet{\mathsfit}{bold}{\encodingdefault}{\sfdefault}{bx}{n}

\usepackage{hyperref}
\usepackage{url}
\usepackage{booktabs} 
\usepackage{graphicx} 
\usepackage{multirow}
\usepackage{pifont}
\usepackage{wrapfig}
\usepackage{caption}
\usepackage{xcolor}
\usepackage{enumitem}

\title{Preemptive Instruction Following with Resumption}

\author{Zerun Liu, Lechao Cheng\thanks{Corresponding Author}, Yimin Liu, Shengeng Tang, Yaxiong Wang, Zhun Zhong \\
School of Computer Science and Information Engineering‌\\
Hefei University of Technology‌\\
Hefei, Anhui 230601, China \\
\texttt{\{liuzerun, chenglc, yiminliu, tangsg, wangyx\}@mail.hfut.edu.cn} \\
\texttt{zhunzhong007@gmail.com}
}

\iclrfinalcopy % Uncomment for camera-ready version, but NOT for submission.
\begin{document}

\maketitle
% Arivx------------------------------------------------------------------------------
\vspace{-2em}
% Arivx------------------------------------------------------------------------------
\begin{abstract}

Long-horizon embodied navigation may be interrupted by a higher-priority instruction before the current task sequence is complete. The agent must finish its current atomic task, execute the inserted task, and resume the remaining sequence from the updated environment state, where its viewpoint and available navigation candidates may have changed. To evaluate this capability, we introduce \textbf{PIFR} benchmark, a novel task for runtime task preemption and resumption in embodied navigation. PIFR includes single, repeated, and nested interruption settings, with stage-specific metrics for inserted-task completion, resumption of the original task, and completion of the full sequence. Evaluations show that recovery performance varies across interruption conditions, while aggregate success can obscure differences in interruption handling and resumption. We further propose a lightweight candidate-ranking baseline to improve target re-grounding after interruption. Together, PIFR and the baseline provide a systematic framework for studying interruption handling and task recovery in long-horizon embodied navigation.

\end{abstract}
\section{Introduction}

Embodied navigation research has advanced the ability of agents to understand their surroundings, follow instructions, and complete navigation tasks through work on vision-language navigation~\citep{anderson2018vision}, embodied simulation~\citep{savva2019habitat}, interactive navigation~\citep{thomason2020vision}, and spatial memory in partially observed environments~\citep{chaplot2020learning,wani2020multion}. Subsequent studies have expanded the range of tasks and goal specifications. For example, TEACh~\citep{padmakumar2022teach} investigates multi-step household tasks guided by human instructions; NaviLLM~\citep{zheng2024towards} models diverse navigation tasks through a unified language interface; and GOAT-Bench~\citep{khanna2024goat} evaluates long-horizon navigation with sequential category, language, and image goals. Other work studies context-dependent sequential grounding~\citep{zhang2024task}, combines visual grounding with active exploration~\citep{zhu2025move}, maintains spatial representations over extended execution~\citep{yang20253d,fan2025embodied}, or investigates asynchronous interaction, task prioritization, and runtime preemption~\citep{wang2025coneqsa,zhou2025p3,chen2026typego}. These studies advance different capabilities relevant to long-horizon navigation, interaction, and task scheduling. However, to the best of our knowledge, existing embodied navigation benchmarks do not systematically evaluate the complete process in which an agent receives a higher-priority instruction while executing a base task sequence, completes its current atomic task before switching to the inserted task, and resumes the base sequence from the updated environmental state after completing the inserted task. Aggregate navigation success alone also cannot distinguish performance in handling the inserted task, resuming the original task, and completing the full sequence. A dedicated benchmark is therefore needed to evaluate runtime task preemption and resumption in a controlled, stage-specific manner.

We study \textit{runtime task preemption and resumption}. Let the base task sequence be \(A=(E_{A_1},E_{A_2},\ldots,E_{A_n})\). If the agent receives a higher-priority instruction while executing an atomic task, it first completes that task, then suspends the remaining tasks in \(A\) and executes the inserted task. Once the inserted task is complete, the agent resumes \(A\) at the next unexecuted atomic task. Here, \textbf{resumption} refers to restoring the logical progress of the base sequence and re-grounding its next objective from the current environmental state; it does not require returning to the pre-interruption position or reusing the original route.
% ---------------------------------ICLR 投稿版本-----------------------------------
\begin{wrapfigure}{R}{7cm}
\centering
\vspace{-4mm}
\includegraphics[width=7cm,
    trim={16cm 0 0 0},
    clip]{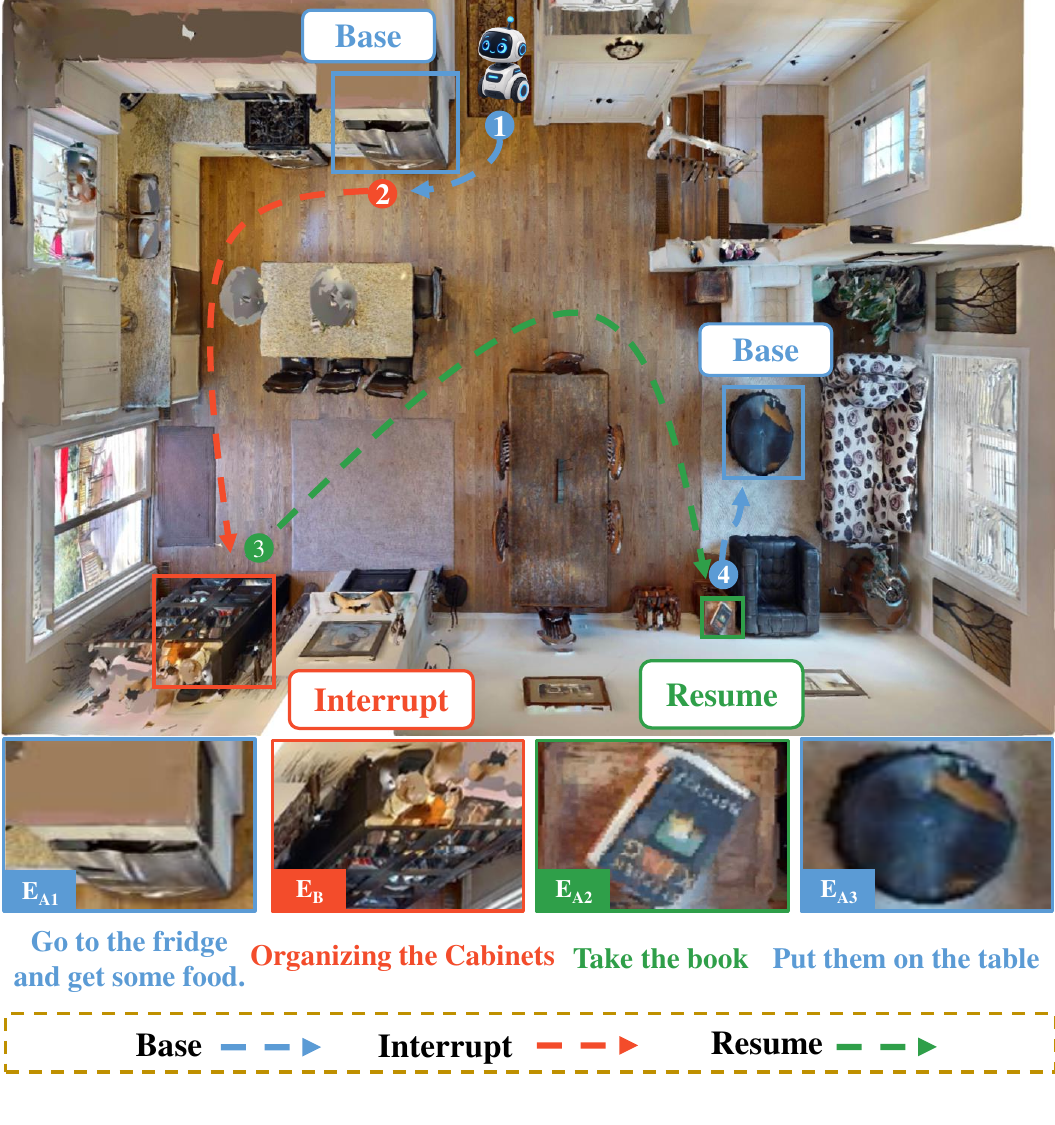}
    \vspace{-8mm}
\caption{\small \textbf{Interruption-and-resumption navigation in PIFR.}
        During long-horizon navigation, an agent may need to suspend its current task, handle an inserted objective, and then return to the unfinished task. Blue, red, and green trajectories indicate base, interrupt, and resume execution, respectively.}
\label{fig:teaser}
\vspace{-6mm}
\end{wrapfigure}
% ---------------------------------ICLR 投稿版本-----------------------------------
% ---------------------------------Arxiv    版本-----------------------------------
% \begin{figure*}[!htb]
% \centering
% \vspace{-4mm}
% \includegraphics[width=\linewidth,height=14cm,
%     trim={16cm 0 0 0},
%     clip]{Figures/PIFR_teaser_v2.pdf}
%     \vspace{-8mm}
% \caption{\small \textbf{Interruption-and-resumption navigation in PIFR.}
%         During long-horizon navigation, an agent may need to suspend its current task, handle an inserted objective, and then return to the unfinished task. Blue, red, and green trajectories indicate base, interrupt, and resume execution, respectively.}
% \label{fig:teaser}
% \vspace{-6mm}
% \end{figure*}
% ---------------------------------Arxiv    版本-----------------------------------
Figure~\ref{fig:teaser} illustrates this process. The agent is initially assigned the base sequence \(A=(E_{A_1},E_{A_2},E_{A_3})\): retrieve food from a refrigerator, retrieve a book, and complete a subsequent task. While executing \(E_{A_1}\), the agent receives a higher-priority inserted task \(E_B\), which requires organizing the cabinets. To ensure that the current atomic task is completed, the agent finishes \(E_{A_1}\) before suspending the unexecuted tasks \(E_{A_2}\) and \(E_{A_3}\) and switching to \(E_B\). After completing \(E_B\), the agent resumes \(A\) at \(E_{A_2}\) and then proceeds to \(E_{A_3}\). Because the inserted task may change the agent's location, viewpoint, and available navigation candidates, resuming \(E_{A_2}\) requires the agent to re-ground its target and plan from the current state, rather than return to its pre-interruption position or follow the original route.

To address this gap, we introduce \textbf{PIFR}, a benchmark for evaluating runtime task preemption and resumption in embodied navigation. PIFR examines whether an agent can handle an inserted task and subsequently resume its suspended task sequence under diverse goal specifications and interruption structures, including \textit{single}, \textit{repeated}, and \textit{nested} interruptions. This design enables evaluation beyond conventional task completion by distinguishing performance during task interruption and subsequent resumption.

Evaluations on PIFR show that recovery performance varies with interruption conditions and goal specifications. They also reveal that aggregate navigation success can mask differences in an agent’s ability to handle inserted tasks and resume suspended ones. To study this challenge, we further propose \textbf{PIFR-Rank}, a lightweight candidate-ranking baseline aimed at improving target re-grounding after an interruption. Together, PIFR and PIFR-Rank provide an evaluation benchmark and a reference method for investigating task recovery in long-horizon embodied navigation.

Our contributions are summarized as follows:
\begin{itemize}
    \item We introduce \textbf{PIFR}, a benchmark for runtime task preemption and resumption in embodied navigation. The curated datasets cover diverse goal specifications and single, repeated, and nested interruption structures.

    \item We develop a continuous-execution protocol and stage-specific evaluation measures to assess inserted-task completion, resumption of the base task, and completion of the full task sequence, enabling systematic analysis of navigation performance under different interruption conditions.

    \item We propose \textbf{PIFR-Rank}, a lightweight candidate-ranking baseline that improves target re-grounding after interruption through candidate-level supervision while preserving the original navigation pipeline.
\end{itemize}

\section{Related Work}
% Our work is closely related to long-horizon embodied navigation and interruption-aware task execution. The former studies sequential goal completion and persistent navigation, while the latter investigates how agents respond to changing instructions during execution.

\subsection{Embodied Navigation and Long-Horizon Task Execution}

Early embodied navigation research develops methods and benchmarks for distinct task formulations, including vision-and-language navigation~\citep{anderson2018vision}, ObjectNav~\citep{chaplot2020object}, and ImageNav~\citep{zhu2017target}. These studies establish foundations for embodied perception and navigation while introducing different goal types and task protocols. Subsequent work explores whether a unified agent can handle a broader range of navigation tasks. Towards Versatile Embodied Navigation~\citep{wang2022towards} learns a unified policy across multiple tasks, while NaviLLM~\citep{zheng2024towards} models diverse embodied tasks through a shared language interface. MTU3D~\citep{zhu2025move} combines visual grounding, active exploration, and spatial memory to support category-, language-, and image-conditioned navigation. OctoNav~\citep{gao2026octonav} further extends generalist navigation to free-form instructions and diverse goal modalities. These studies primarily broaden the types of tasks and goals that an agent can handle.

A complementary line of work investigates multi-goal navigation and long-horizon task execution. MultiON~\citep{wani2020multion} studies whether an agent can visit multiple targets sequentially within a single episode. GOAT-Bench~\citep{khanna2024goat} evaluates long-term navigation with goals specified as categories, language, or images. Long-Horizon Vision-Language Navigation~\citep{song2025towards} considers multi-stage instructions and consecutive subtask execution, while Lifelong Embodied Navigation Learning~\citep{wang2026lifelong} studies the acquisition of new navigation capabilities from a continual-learning perspective. Overall, these works extend the scale, duration, or goal diversity of navigation tasks, but primarily evaluate the execution of predefined tasks or task sequences. They do not systematically examine a different execution scenario in which a new task is inserted before the original sequence is complete and the agent must resume that sequence from the environment state reached after executing the inserted task.

%\subsection{Task Interruption and Resumption}

\subsection{Interruptions in Embodied Task Execution}

Embodied agents often receive new instructions during ongoing interactions. HANNA~\citep{nguyen2019help} enables navigation agents to request multimodal assistance from a simulated human, while Vision-and-Dialog Navigation~\citep{thomason2020vision} and TEACh~\citep{padmakumar2022teach} study instruction-driven interaction and multi-step household task execution. These works examine how agents use additional instructions or dialogue to guide ongoing behavior, but do not evaluate the recovery of an unfinished navigation task after the agent executes an intervening task. Recent proposed InterruptBench~\citep{zou2026users} explicitly studies user interruptions during web-agent execution, including changes to ongoing requests. Its setting concerns modifying the active user request. In this work, however, PIFR instead considers embodied navigation, where an inserted task is executed while the original task sequence remains unfinished and must be resumed from the resulting environmental state. This setting requires evaluating both the execution of the inserted task and the recovery of the suspended navigation task. To the best of our knowledge, existing embodied navigation benchmarks do not provide a systematic evaluation of this process.

\section{PIFR Benchmark}
\label{sec:pifr_benchmark}

We introduce \textbf{PIFR}, a novel task for evaluating interruption and resumption in persistent embodied navigation. Unlike conventional sequential navigation, in which an agent proceeds through a predefined sequence of objectives, PIFR allows an unfinished task sequence to be suspended by an inserted task and resumed after that task is completed. We first formally define the PIFR task (sec.~\ref{sec:pifr_task}) and present three representative instantiations (sec.~\ref{sec:inst}). We then describe how the datasets are constructed under this formulation and provide the corresponding implementation details (sec.~\ref{sec:data_curation}).

% The environment is not reset during this process. 
%PIFR is constructed from SG3D~\citep{zhang2024task}, GOAT-Bench~\citep{khanna2024goat}, and HM3D-OVON~\citep{yokoyama2024hm3d}. It preserves their respective goal specifications while introducing a unified protocol for task interruption and resumption.

\subsection{Task Definition}
\label{sec:pifr_task}

Let $\mathcal{A}$ denote the set of atomic navigation tasks. Each task $x\in\mathcal{A}$ specifies a navigation goal and its corresponding instruction. A base task sequence is a finite ordered sequence
\[
X=(x_1,x_2,\ldots,x_n), \qquad x_i\in\mathcal{A}.
\]
Let $\mathcal{T}_0=\mathcal{A}^{+}$ denote the set of all nonempty finite sequences of atomic tasks. We use $X[i{:}j]$ to denote the contiguous subsequence of $X$ from task $i$ to task $j$.

We recursively define composite task sequences to represent interruptions. Let $\mathcal{K}(X)\subseteq\{1,\ldots,n-1\}$ denote the set of valid interruption boundaries in $X$. An interruption at boundary $k$ occurs after $x_k$ is completed and before $x_{k+1}$ begins. Given two task sequences or composite task sequences $X$ and $Y$, embedding $Y$ into $X$ at boundary $k\in\mathcal{K}(X)$ is denoted by $X\oplus_kY$ and defined as
\[
X\oplus_kY
:=
X[1{:}k]\rightarrow Y\rightarrow X[k+1{:}n],
\]
where $\rightarrow$ denotes execution order. The embedded sequence $Y$ may itself be a composite sequence, allowing interruptions to be recursively nested.

The set of composite tasks is defined recursively as
\[
\mathcal{T}_{d+1}
=
\mathcal{T}_d
\cup
\left\{
X\oplus_kY
\;\middle|\;
X,Y\in\mathcal{T}_d,\ k\in\mathcal{K}(X)
\right\},
\qquad d\geq 0,
\]
where $\mathcal{T}_d$ contains task sequences with nesting depth at most $d$. The set of all finite composite task sequences is
\[
\mathcal{T}_{\mathrm{comp}}
=
\bigcup_{d=0}^{\infty}\mathcal{T}_d.
\]
Each PIFR event is an instance in $\mathcal{T}_{\mathrm{comp}}$, subject to the interruption and nesting constraints specified by the benchmark.

During a composite event, the simulator state is preserved across interruptions and resumptions. Once an inserted sequence is completed, the suspended sequence resumes at its next unfinished task from the resulting environment state. Thus, resumption preserves task progress but does not restore the agent's pre-interruption position or trajectory. The instruction for the next unfinished task is re-issued upon resumption, requiring the agent to ground its goal in the updated physical and perceptual context.

\subsection{Canonical Instantiations}\label{sec:inst}

Let $A=(a_1,\ldots,a_m)$ be a base sequence, and let $B$ and $C$ be inserted sequences. The following three PIFR settings are instances of the recursive embedding definition.

\ding{182}~\textbf{Single interruption.}
A single inserted sequence $B$ is embedded at boundary $k_A$ of $A$:
\[
A\oplus_{k_A}B
=
A[1{:}k_A]\rightarrow B\rightarrow A[k_A+1{:}m].
\]
The agent executes the prefix of $A$, then $B$, and finally resumes $A$ at its next unfinished task.

\ding{183}~\textbf{Repeated interruption.}
Two inserted sequences, $B$ and $C$, are embedded successively at distinct boundaries $k_1<k_2$ of the base sequence $A$:
\[
\bigl(A\oplus_{k_1}B\bigr)\oplus_{k_2}C
=
A[1{:}k_1]\rightarrow B
\rightarrow A[k_1+1{:}k_2]
\rightarrow C
\rightarrow A[k_2+1{:}m].
\]
Here, $k_1$ and $k_2$ are indexed with respect to the original sequence $A$. After completing $B$, the agent resumes $A$ and continues to the second interruption boundary. It then executes $C$ before resuming the remaining suffix of $A$.

\ding{184}~\textbf{Nested interruption.}
Sequence $B$ is embedded in $A$ at boundary $k_A$, and sequence $C$ is embedded in $B$ at boundary $k_B$:
\[
A\oplus_{k_A}\bigl(B\oplus_{k_B}C\bigr)
=
A[1{:}k_A]\rightarrow B[1{:}k_B]
\rightarrow C
\rightarrow B[k_B+1{:}n]
\rightarrow A[k_A+1{:}m],
\]
where $B=(b_1,\ldots,b_n)$. The agent first executes the prefix of $A$ and then the prefix of $B$. Sequence $B$ is subsequently interrupted by $C$. After completing $C$, the agent resumes and completes $B$, and then resumes $A$. This defines the last-in, first-out resumption order induced by recursive embedding.

\begin{figure*}
    \centering
    \includegraphics[width=\textwidth,
        trim={0 2.8cm 0 5.8cm}, % 左15pt，下10pt，右15pt，上10pt
        clip]{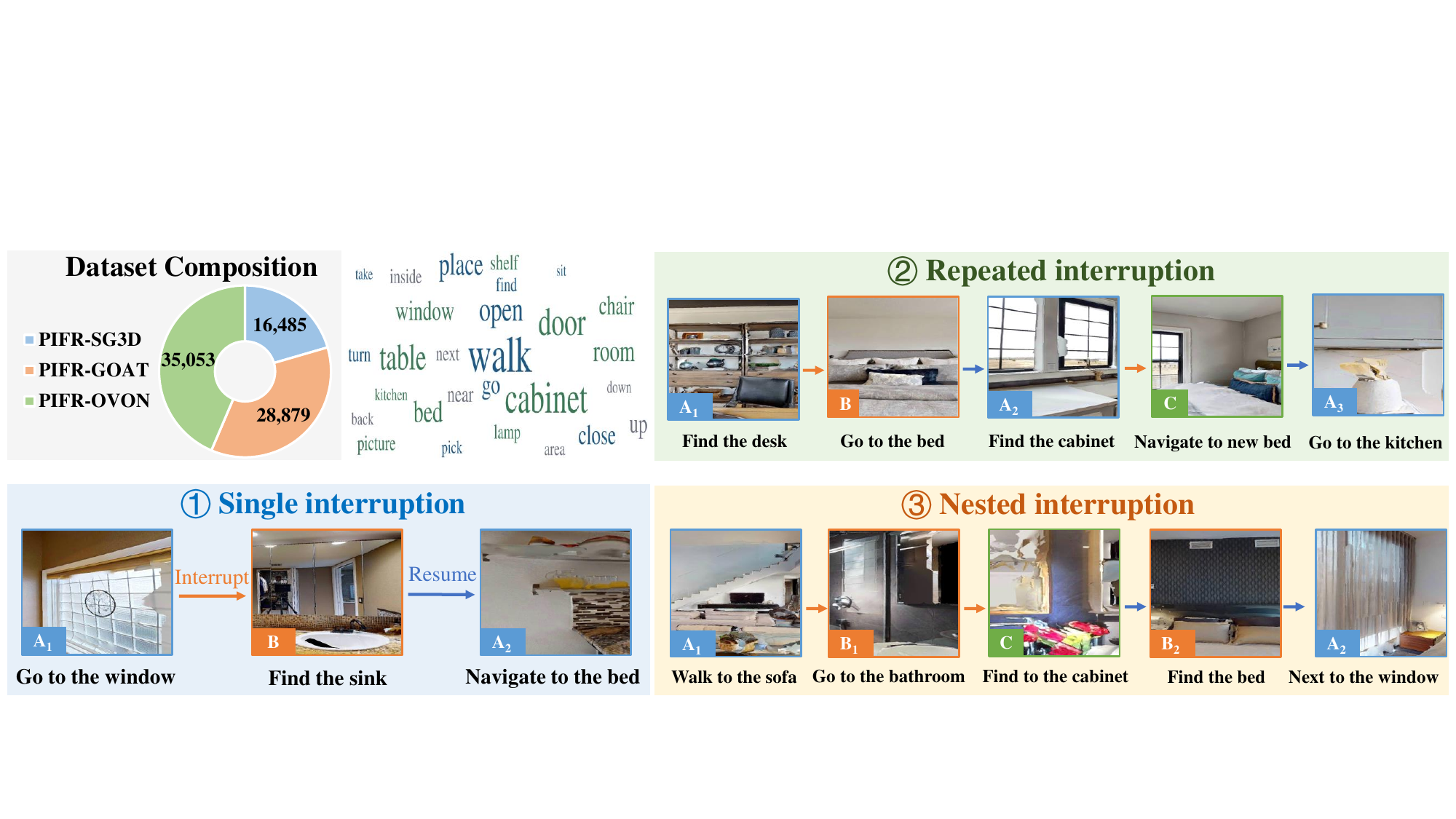}
    \caption{\small
        \textbf{The overview of the composition of PIFR and the Single, Repeated and Nested interruption.} Among the three different types of interrupts—Single, Repeated, and Nested—the orange arrow represents an interrupt, and the blue arrow represents a resume.
    }
    \label{fig:pifr_construction}
    \vspace{-5mm}
\end{figure*}

As shown in Figure~\ref{fig:pifr_construction}, The three settings differ in their embedding structure: \textbf{Single} embeds one episode at one boundary; \textbf{Repeated} embeds episodes at multiple boundaries of the same base episode; and \textbf{Nested} recursively embeds an interruption within an inserted episode. Any limits on the number of interruptions or nesting depth are specified by the benchmark construction protocol.

\subsection{Dataset Curation} \label{sec:data_curation}
We construct three PIFR dataset variants from SG3D~\citep{zhang2024task}, GOAT-Bench~\citep{khanna2024goat}, and HM3D-OVON~\citep{yokoyama2024hm3d}: \textbf{PIFR-SG3D}, \textbf{PIFR-GOAT}, and \textbf{PIFR-OVON}, respectively. Each variant inherits the goal specifications of its source benchmark. Specifically, PIFR-SG3D supports description-based instance grounding, PIFR-GOAT includes category, language, and image goals, and PIFR-OVON supports open-vocabulary object-category navigation. We preserve these goal specifications and compose the source navigation tasks into the composite task episodes defined above, enabling the evaluation of task resumption across diverse goal settings.

\paragraph{Data Splits and Statistics.}
We retain the original data splits of each source benchmark. PIFR-SG3D uses the \texttt{train} and \texttt{val} splits, whereas PIFR-GOAT and PIFR-OVON use the \texttt{train}, \texttt{val\_seen}, \texttt{val\_seen\_synonyms}, and \texttt{val\_unseen} splits. As reported in Table~\ref{tab:pifr_statistics}, PIFR contains 80,417 composite episodes, comprising 74,223 training episodes and 6,194 validation episodes. Specifically, PIFR-SG3D contains 16,019 training and 466 validation episodes; PIFR-GOAT contains 28,148 training and 731 validation episodes; and PIFR-OVON contains 30,056 training and 4,997 validation episodes.

The training set contains 59,101 Single episodes (79.63\%), 7,558 Repeated episodes (10.18\%), and 7,564 Nested episodes (10.19\%), yielding an approximate distribution of 80\%/10\%/10\%. Each composite episodes comprises multiple atomic navigation tasks. Across the full benchmark, an episodes contains 9.78 atomic navigation segments on average, with a median of 8. The mean episodes lengths are 9.23, 11.62, and 12.30 segments for the Single, Repeated, and Nested settings, respectively.

\paragraph{Episode Construction and Validation.}
For PIFR-SG3D and PIFR-GOAT, we select base episodes from source episodes containing at least four atomic navigation tasks, thereby providing multiple valid embedding boundaries. Since HM3D-OVON consists of single-goal episodes, we deterministically combine four episodes from the same scene, each with a different target category, to form a multi-task base episode. Inserted episodes are sampled from compatible episodes in the same scene, and their first goals are required to differ from those in the parent episode.

We construct composite events by embedding inserted episodes at valid atomic-task boundaries according to the previously defined $\oplus_k$ operation. The Single setting samples one embedding boundary; the Repeated setting samples two distinct boundaries in the base episode; and the Nested setting additionally requires a valid boundary within an inserted episode, allowing another episode to be embedded recursively. Because individually feasible source episodes may become infeasible when combined, we verify geometric navigability along the complete execution chain. We also use deterministic sampling and balance events across interruption structures and scenes to support reproducibility and dataset diversity.

\begin{table*}[!hb]
\centering
\begin{minipage}[t]{0.49\textwidth}
\vspace{0pt}
\centering
\captionof{table}{
\small  PIFR dataset.
\textbf{Seen} means categories and instances are in the training data.
\textbf{Syn.} denotes unseen descriptions use synonyms of goals observed during training.
\textbf{Unseen} indicates categories are not in the training data.
}
\label{tab:pifr_statistics}
\vspace{-3mm}
\renewcommand{\arraystretch}{0.9}
\resizebox{\linewidth}{!}{
\begin{tabular}{lrrrrr}
\toprule
% \textbf{Variant} & \textbf{Train} & \textbf{Val/Seen} & \textbf{Syn.} & \textbf{Unseen} & \textbf{Total} \\
\multirow{2}{*}{\textbf{Variant}}
& \multirow{2}{*}{\textbf{Train}}
& \multicolumn{3}{c}{\textbf{Test}}
& \multirow{2}{*}{\textbf{Total}} \\
\cmidrule(lr){3-5}
& & \textbf{Seen} & \textbf{Syn.}
& \textbf{Unseen} & \\
\midrule

PIFR-SG3D
& 16,019 & 466 & -- & -- & 16,485 \\

PIFR-GOAT
& 28,148 & 239 & 248 & 244 & 28,879 \\

PIFR-OVON
& 30,056 & 1,631 & 1,623 & 1,743 & 35,053 \\

\midrule

\textbf{Total}
& \textbf{74,223}
& \multicolumn{3}{c}{\textbf{6,194}}
& \textbf{80,417} \\

\bottomrule
\end{tabular}
}
\end{minipage}
\hfill
% ==================== Left Table ====================
\begin{minipage}[t]{0.49\textwidth}
\vspace{0pt}
\centering
\captionof{table}{
\small Comparison with representative embodied navigation benchmarks.
\textbf{Pers.}, \textbf{Int.}, and \textbf{Res.} denote persistent execution,
interruption, and resumption.
}
\label{tab:benchmark_comparison}
\vspace{-2mm}
\renewcommand{\arraystretch}{1.5}
\resizebox{\linewidth}{!}{
\begin{tabular}{lccccc}
\toprule
\textbf{Benchmark} &
\textbf{Descriptio}n &
\textbf{Pers.} &
\textbf{Int.} &
\textbf{Res.} & \\
\midrule

HM3D-OVON
& $A$
& $\textcolor{red}{\times}$
& $\textcolor{red}{\times}$
& $\textcolor{red}{\times}$ \\

MultiON
& $A \rightarrow B \rightarrow C$
& $\textcolor{green}{\checkmark}$
& $\textcolor{red}{\times}$
& $\textcolor{red}{\times}$ \\

GOAT-Bench
& $A \rightarrow B \rightarrow C$
& $\textcolor{green}{\checkmark}$
& $\textcolor{red}{\times}$
& $\textcolor{red}{\times}$ \\

LHPR-VLN
& $A \rightarrow B \rightarrow C$
& $\textcolor{green}{\checkmark}$
& $\textcolor{red}{\times}$
& $\textcolor{red}{\times}$ \\

\midrule

\textbf{PIFR}
& \textbf{Single / Repeated / Nested}
& $\textcolor{green}{\mathbf{\checkmark}}$
& $\textcolor{green}{\mathbf{\checkmark}}$
& $\textcolor{green}{\mathbf{\checkmark}}$ \\

\bottomrule
\end{tabular}
}
\end{minipage}

% ==================== Right Table ====================
\vspace{-5mm}
\end{table*}

\paragraph{Comparison with Existing Benchmarks.}
Table~\ref{tab:benchmark_comparison} compares PIFR with representative embodied navigation benchmarks in terms of temporal task organization. Existing benchmarks cover single-goal navigation, ordered multi-goal tasks, and long-horizon execution, but generally require agents to progress through objectives in a predefined order. PIFR additionally supports interruption and resumption of task sequences through recursive embedding. Its Nested setting further evaluates recovery when an inserted sequence is itself interrupted.

\section{PIFR-Rank}
\label{sec:method}
\begin{wrapfigure}{R}{7cm}
\centering
\vspace{-12mm}
\includegraphics[width=7cm,
    trim={14.5cm 0 0 0},
    clip]{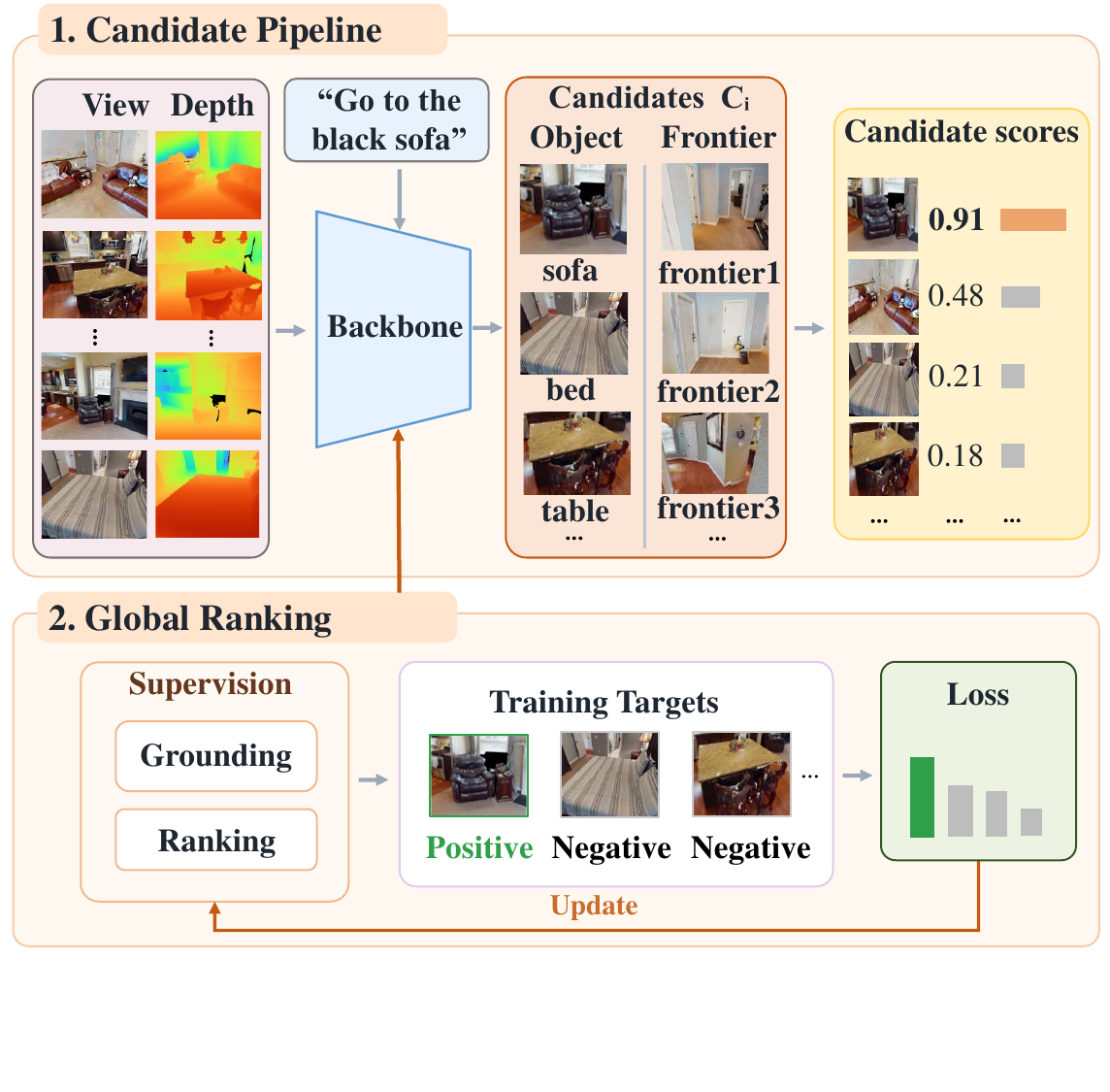}
\vspace{-14mm}
\caption{\small Overview of PIFR-Rank. The upper shows object and frontier candidate generation from visual observations and the active instruction. The lower illustrates the ranking objective, which encourages target-consistent candidates to score above distractors.}
\label{fig:method}
%\vspace{5mm}
\end{wrapfigure}

\paragraph{Overview.}
When an inserted task sequence is completed, the agent resumes the most recently suspended sequence from the resulting environment state and receives its next unfinished atomic-task instruction. Since the intervening execution may change the agent's location, observations, and spatial context, the goal must be re-grounded among the currently available candidates. To improve this process, we propose \textbf{PIFR-Rank}, which trains the grounding scorer with candidate-level supervision and a ranking loss, while assigning greater weight to decisions made upon resumption. Figure~\ref{fig:method} illustrates the candidate-generation pipeline and the proposed ranking objective.

\subsection{Candidate Construction and Training Objective}
\label{sec:candidate_objective}

At grounding decision $i$, the system constructs a candidate set
\(
\mathcal{C}_i=\{c_{ij}\}_{j=1}^{N_i}
\)
from the current observation and navigation context. It includes object candidates from observed regions and frontier candidates associated with unexplored regions. Given the active instruction $q_i$, the grounding scorer assigns each candidate a score
\[
s_{ij}=f_{\theta}(c_{ij},q_i),
\]
where $f_{\theta}$ denotes the scorer optimized during training.

Each candidate receives a binary label $y_{ij}\in\{0,1\}$ indicating whether it is valid for the active goal. When the target has been observed, matching object candidates are labeled positive; if multiple candidates match, all receive positive labels. When the target has not been observed, training-only target and scene information is used to label a suitable frontier candidate from the existing candidate set. These labels do not add candidates or affect candidate generation or selection at inference time.

Let $m_{ij}\in\{0,1\}$ indicate whether an entry is a valid candidate ($m_{ij}=1$) or padding ($m_{ij}=0$). For each decision with at least one valid candidate and one positive candidate, the padding-aware binary cross-entropy loss is
\begin{equation}
\ell_i^{\mathrm{BCE}}
=
\frac{
\sum_j m_{ij}\,
\operatorname{BCEWithLogits}(s_{ij},y_{ij})
}{
\sum_j m_{ij}
}.
\end{equation}
This loss supports multiple positive candidates and excludes padding entries.

To encourage positive candidates to outrank distractors, we define
\begin{equation}
s_i^{+}
=
\max_{j:\,m_{ij}=1,\,y_{ij}=1}s_{ij},
\qquad
s_i^{-}
=
\max_{j:\,m_{ij}=1,\,y_{ij}=0}s_{ij}.
\end{equation}
When both positive and negative candidates are present, the ranking loss is
\begin{equation}
\ell_i^{\mathrm{rank}}
=
\max\left(0,\,\delta-s_i^{+}+s_i^{-}\right),
\end{equation}
where $\delta>0$ is the margin. It requires the highest-scoring positive to exceed the highest-scoring negative by at least $\delta$; it does not constrain every positive-negative pair. The ranking loss is zero when no valid negative candidate is available. Decisions without a valid positive candidate are excluded from both candidate-level losses.

A \emph{resumption decision} is the grounding decision for the next unfinished atomic task when a suspended sequence resumes after an inserted sequence completes. This definition applies to each resumption in Repeated and Nested events, including the resumption of an inner sequence and of the outer base sequence. Let $r_i$ denote the execution role of decision $i$, and let $w(r_i)$ assign greater weight to resumption decisions. For the set $\mathcal{I}$ of decisions with at least one valid candidate and one positive candidate, we optimize
\begin{equation}
\mathcal{L}_{\mathrm{GR}}
=
\frac{
\sum_{i\in\mathcal{I}} w(r_i)
\left(
\ell_i^{\mathrm{BCE}}+\ell_i^{\mathrm{rank}}
\right)
}{
\sum_{i\in\mathcal{I}} w(r_i)
}.
\end{equation}
Thus, decisions with positive candidates but no negative candidates still contribute their BCE loss. Execution-role labels affect only training weights and are not inputs to the online policy.

% The final objective is
% \begin{equation}
% \mathcal{L}
% =
% \mathcal{L}_{\mathrm{task}}
% +
% \lambda_{\mathrm{GR}}\mathcal{L}_{\mathrm{GR}},
% \end{equation}
% where $\mathcal{L}_{\mathrm{task}}$ is the task-training objective and $\lambda_{\mathrm{GR}}$ controls the contribution of candidate-level supervision.

% \subsection{Inference}
% \label{sec:inference}
At inference time, the agent uses the active instruction, visual observations, navigation context, and candidates generated by the navigation system. Candidate labels, execution-role annotations, future tasks, and training-only reference information are unavailable. PIFR-Rank changes the training objective while retaining the original candidate-generation and online navigation procedures.

\section{experiments}
\label{sec:experiments}
% \subsection{Experimental Setup}
% \label{sec:experimental_setup}
\paragraph{Implementation Details.}
PIFR-Rank is fine-tuned independently on PIFR-SG3D, PIFR-GOAT, and PIFR-OVON. We optimize the model for 5 epochs using AdamW with a learning rate of $1\times10^{-4}$, $\beta=(0.9,0.98)$, and weight decay of $0.01$. The GlobalRank objective uses a ranking margin $\delta=0.5$, and a Resume sample weight of $3$. All models are trained on NVIDIA RTX 4090 GPUs with a batch size of 8. 
\paragraph{Metrics.}
Following standard embodied navigation evaluation~\citep{khanna2024goat}, we report \textbf{Success Rate (SR)} and \textbf{Success weighted by Path Length (SPL)} to measure navigation success and path efficiency. An atomic navigation segment is considered successful if the agent terminates within 0.25m geodesic distance of a valid target viewpoint. SR and SPL are computed over completed atomic navigation segments. To evaluate interruption-and-resumption behavior, we additionally report \textbf{Interrupt SR} and \textbf{Resume SR}. For each segment $q$, let $d_q$ and $d_q^{\mathrm{end}}$ denote the shortest path length and the geodesic distances to the active target at its end, respectively, and let $\ell_q$ be the agent path length, $\mathcal{Q}$ denote the set of valid atomic navigation segments. A segment is successful if $S_q=[d_q^{\mathrm{end}}\leq\tau]$, where $\tau=0.25\,\mathrm{m}$. The four metrics are shown below:

% Let $\mathcal{T}_{\mathrm{int}}$ and $\mathcal{T}_{\mathrm{res}}$ denote the sets of interruption and resumption segments, and let $S_q\in\{0,1\}$ indicate whether segment $q$ succeeds. We define $\mathrm{Interrupt\ SR} = \frac{\sum_{q\in\mathcal{T}_{\mathrm{int}}} S_q} {|\mathcal{T}_{\mathrm{int}}|}\times 100\%$, $\mathrm{Resume\ SR} = \frac{\sum_{q\in\mathcal{T}_{\mathrm{res}}} S_q} {|\mathcal{T}_{\mathrm{res}}|}\times 100\%.$ They measure the success rates of navigation segments associated with interruption and resumption.

\begin{itemize}[
    leftmargin=0pt,
    labelsep=0.45em,
    itemsep=1pt,
    topsep=2pt,
    parsep=0pt,
    partopsep=0pt
]
\setlength{\abovedisplayskip}{2pt}
\setlength{\belowdisplayskip}{2pt}

    \item \textbf{Overall navigation metrics.}
    Success Rate (\textbf{SR}) measures the proportion of successful atomic navigation segments, while Success weighted by Path Length (\textbf{SPL}) accounts for path efficiency:
    \begin{equation}
    \begin{aligned}
        \mathrm{\textbf{SR}}
        &:= \frac{100}{|\mathcal{Q}|}
        \sum_{q\in\mathcal{Q}} S_q,
        \qquad
        \mathrm{\textbf{SPL}}
        := \frac{100}{|\mathcal{Q}|}
        \sum_{q\in\mathcal{Q}}
        S_q
        \frac{d_q}
        {\max(d_q,\ell_q,\epsilon)}.
    \end{aligned}
    \end{equation}

    \item \textbf{Stage-specific metrics.}
    Interruption SR (\textbf{Int.}) and Resumption SR (\textbf{Res.}) measure the success rates of interruption-entry and resumption-entry segments, respectively. Let $\mathcal{T}_{\mathrm{int}}$ and $\mathcal{T}_{\mathrm{res}}$ denote the corresponding sets of valid segments:
    \begin{equation}
    \begin{aligned}
        \mathrm{\textbf{Int.}}
        &:= \frac{100}{|\mathcal{T}_{\mathrm{int}}|}
        \sum_{q\in\mathcal{T}_{\mathrm{int}}} S_q,
        \qquad
        \mathrm{\textbf{Res.}}
        := \frac{100}{|\mathcal{T}_{\mathrm{res}}|}
        \sum_{q\in\mathcal{T}_{\mathrm{res}}} S_q.
    \end{aligned}
    \end{equation}
\end{itemize}

\subsection{Benchmarking results with Navigation Baselines}
\label{sec:main_results}

% \subsection{Benchmark Results}
% \label{sec:main_results}
%----------------------------------ICLR 版本----------------------------------
\begin{table*}
%----------------------------------ICLR 版本----------------------------------
%----------------------------------Arxiv 版本----------------------------------
% \begin{table*}[!htb]
%----------------------------------Arxiv 版本----------------------------------
\centering
\vspace{-3mm}
\caption{\small Results on PIFR-SG3D / PIFR-OVON (Seen) / PIFR-GOAT (Seen) under different interruption settings (\textbf{\ding{182} Single Interruption, \ding{183} Repeated Interruption, \ding{184} Nested Interruption} ).}
\label{tab:pifr_bench}
\vspace{-3mm}
%\begingroup
% \footnotesize
%\scriptsize
\renewcommand{\arraystretch}{0.8}
% \setlength{\tabcolsep}{2.4pt}
% \renewcommand{\arraystretch}{1.05}
% \begin{tabular*}{\linewidth}
% {@{}l|c@{\extracolsep{\fill}}ccc|cccc|cccc@{}}
% \begin{tabular}{@{}l*{12}{c}@{}}
% \begin{tabular}{@{}l|c@{\extracolsep{\fill}}ccc|cccc|cccc@{}}
\resizebox{0.9\textwidth}{!}{
\begin{tabular}{l|cccc|cccc|cccc}
\midrule
\multirow{2}{*}{Method}
& \multicolumn{4}{c|}{\textbf{\ding{182} Single Interruption}}
& \multicolumn{4}{c|}{\textbf{\ding{183}  Repeated Interruption}}
& \multicolumn{4}{c}{\textbf{\ding{184} Nested Interruption}} \\
\cmidrule{2-13}
& SR$\uparrow$ & SPL$\uparrow$ & Int.$\uparrow$ & Res.$\uparrow$
& SR$\uparrow$ & SPL$\uparrow$ & Int.$\uparrow$ & Res.$\uparrow$
& SR$\uparrow$ & SPL$\uparrow$ & Int.$\uparrow$ & Res.$\uparrow$ \\
\midrule
\multicolumn{13}{c}{\textbf{PIFR-SG3D}} \\
\cmidrule{2-13}
R2F
& 4.8 & 2.8 & 3.2 & 3.0
& 3.7 & 1.9 & 3.5 & 5.0
& 3.3 & 2.3 & 3.1 & 2.8 \\
E-VA
& 4.7 & 3.5 & 3.2 & 6.3
& 4.9 & 3.4 & 2.5 & 10.0
& 6.5 & 4.7 & 5.7 & 10.4 \\
Uni-NaVid
& 10.7 & 7.1 & 9.0 & 9.0
& 7.0 & 4.6 & 7.4 & 7.5
& 6.0 & 3.6 & 4.6 & 3.8 \\
MTU3D
& 22.6 & 12.8 & 22.0 & 23.0
& 15.8 & \textbf{11.4} & 11.3 & 20.0
& 15.6 & 9.2 & \textbf{11.3} & \textbf{15.1} \\
\cmidrule{2-13}
% \midrule
\textbf{Ours}
& \textbf{26.7} & \textbf{15.5} & \textbf{22.5} & \textbf{31.4}
& \textbf{17.4} & 11.1 & \textbf{12.5} & \textbf{25.0}
& \textbf{18.9} & \textbf{11.3} & \textbf{11.3} & 13.2 \\
\midrule
\multicolumn{13}{c}{\textbf{PIFR-OVON}} \\
\cmidrule{2-13}
R2F
& 24.3 & 10.4 & 24.4 & 25.5
& 22.4 & 10.2 & 23.5 & 19.1
& 23.2 & 10.2 & 19.3 & 22.3 \\

E-VA
& 20.4 & 10.2 & 22.6 & 13.7
& 19.2 & 10.3 & 20.5 & 17.6
& 21.6 & 11.4 & 22.0 & 21.5 \\

Uni-NaVid
& 34.7 & 13.0 & 37.8 & 31.6
& 34.6 & 12.9 & 29.9 & 32.4
& 33.2 & 12.1 & 31.7 & 32.2 \\

MTU3D
& 49.6 & 23.7 & 44.9 & \textbf{56.1}
& 45.9 & 22.0 & 40.7 & 45.6
& 46.0 & 23.2 & 41.1 & 45.1 \\
\cmidrule{2-13}
\textbf{Ours}

& \textbf{51.7} & \textbf{26.5} & \textbf{49.0} & 49.0
& \textbf{47.5} & \textbf{23.9} & \textbf{47.2} & \textbf{46.2}
& \textbf{47.3} & \textbf{24.9} & \textbf{43.8} & \textbf{50.0} \\
\midrule
\multicolumn{13}{c}{\textbf{PIFR-GOAT}} \\
\cmidrule{2-13}
R2F
& 17.5 & 6.6 & 15.6 & 15.8
& 15.4 & 7.5 & 13.5 & 14.0
& 18.8 & 8.6 & 15.8 & 24.0 \\

E-VA
& 15.4 & 11.3 & 15.5 & 17.5
& 16.9 & 11.1 & 18.0 & 20.0
& 16.7 & 11.6 & 8.0 & 24.0 \\
Uni-NaVid
& 23.8 & 12.2 & 19.5 & 24.2
& 19.6 & 10.3 & 16.5 & 19.0
& 22.6 & 11.1 & 15.4 & 26.0\\
MTU3D
& \textbf{52.8} & \textbf{32.6} & \textbf{47.4} & \textbf{61.1}
& 42.9 & 28.1 & 36.0 & 54.0
& 44.3 & 29.9 & 34.0 & 48.0 \\
\cmidrule{2-13}
\textbf{Ours}
& 50.0 & 31.7 & 44.9 & 58.2
& \textbf{43.8} & \textbf{29.7} & \textbf{38.0} & \textbf{55.0}
& \textbf{49.6} & \textbf{34.8} & \textbf{48.0} & \textbf{60.0}\\
\midrule
\end{tabular}}
%\vspace{-5mm}
%\endgroup
\end{table*}

We evaluate R2F~\citep{argenziano2026r2f}, E-VA~\citep{fan2025embodied}, Uni-NaVid~\citep{zhang2024uni}, MTU3D~\citep{zhu2025move} , and PIFR-Rank on PIFR-SG3D, PIFR-GOAT, and PIFR-OVON under the Single, Repeated, and Nested interruption settings.

\paragraph{Main Results on Seen Goals}
Table~\ref{tab:pifr_bench} compares representative navigation methods on the Seen splits of PIFR-SG3D, PIFR-OVON, and PIFR-GOAT under single, repeated, and nested interruption. The results show that method performance depends on both the source benchmark and the interruption structure. PIFR-Rank achieves higher SR than MTU3D across all three structures on PIFR-SG3D and PIFR-OVON. On PIFR-GOAT, MTU3D performs better under single interruption, whereas PIFR-Rank achieves higher SR under repeated and nested interruption. In particular, under nested interruption, PIFR-Rank improves SR from 44.3\% to 49.6\%.

The stage-specific metrics provide additional information beyond overall navigation success. For example, on PIFR-OVON under single interruption, PIFR-Rank obtains a higher SR than MTU3D (51.7\% vs.\ 49.6\%), but a lower Resume SR (49.0\% vs.\ 56.1\%). This difference shows that overall task success and recovery performance do not always change together, motivating the separate analysis of goal-split generalization and multi-round recovery below.

% Tables~\ref{tab:pifr_bench} and~\ref{tab:pifr_ovon} compare representative methods across goal specifications and interruption structures. No method performs uniformly best across all settings: for example, MTU3D achieves higher success on PIFR-GOAT Seen under Single interruption, whereas PIFR-Rank performs better under Repeated and Nested interruptions. On PIFR-OVON Unseen, PIFR-Rank improves success over MTU3D across all three structures. These results show that relative performance depends on both the goal setting and interruption structure.

\paragraph{Results on Synonym and Unseen Goals}
\label{sec:goal_split_results}

\begin{figure*}[!b]
\vspace{-3mm}
    \centering
    \includegraphics[width=\textwidth]{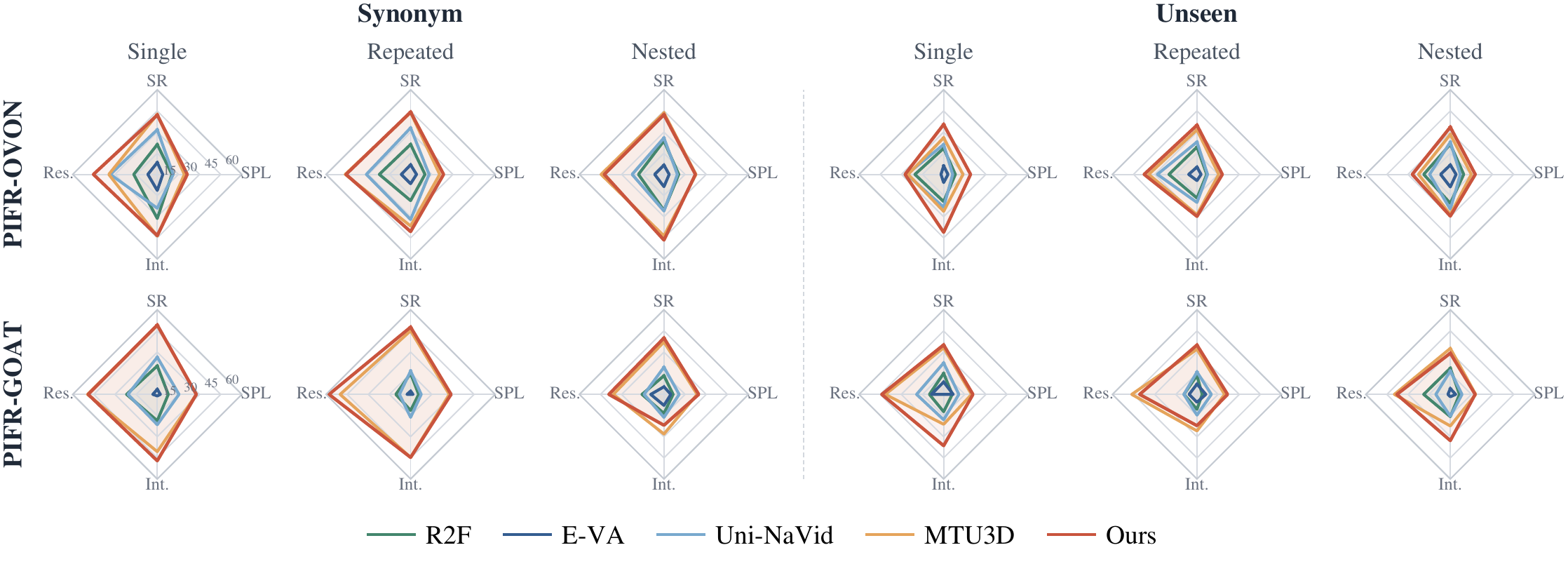}
    \vspace{-6mm}
 \caption{\small Radar plots of SR, SPL, Interrupt SR, and Resume SR on the Synonym and Unseen splits of PIFR-OVON and PIFR-GOAT under single, repeated, and nested interruption.}
    \label{fig:goal_split_radar}
    %\vspace{-3mm}
\end{figure*}

Figure~\ref{fig:goal_split_radar} presents SR, SPL, Interrupt SR, and Resume SR on the Synonym and Unseen splits of PIFR-OVON and PIFR-GOAT. The results show that performance changes are not uniform across datasets or interruption structures. On PIFR-OVON, PIFR-Rank achieves higher SR than MTU3D under all three interruption structures on the Unseen split, whereas its advantage on the Synonym split is small and varies by structure. On PIFR-GOAT, PIFR-Rank outperforms MTU3D on the Synonym split and under single and repeated interruption on the Unseen split, but not under nested interruption. These results indicate that robustness to synonym variation and generalization to unseen goals present different challenges, and that their effects depend on the temporal structure of the episode. The four metrics also reveal that changes in overall success do not necessarily imply corresponding changes in interruption handling or resumption. Since the source benchmarks define their Synonym and Unseen splits independently, comparisons should be made within each benchmark and split protocol.

\paragraph{Stage-wise Recovery under Multiple Interruptions}
\label{sec:multi_round_recovery}

\begin{figure*}
\centering
\vspace{-5mm}
    \includegraphics[width=\textwidth]{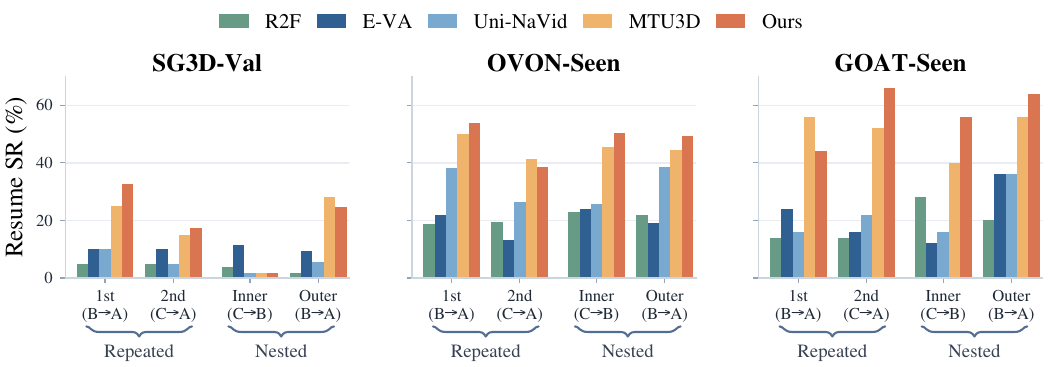}
    \vspace{-6mm}
    \caption{\small
        \textbf{Repeated and nested execution reveal distinct stage-wise recovery profiles.}
        Resume SR is decomposed into the first and second resumptions for ABACA
        ($B\!\rightarrow\!A$ and $C\!\rightarrow\!A$), and the inner and outer resumptions for ABCBA
        ($C\!\rightarrow\!B$ and $B\!\rightarrow\!A$). Recovery behavior differs substantially across navigation settings, showing that a single ABA recovery score cannot fully characterize repeated and nested task resumption.
        }
        \vspace{-5mm}
    \label{fig:stage_recovery}
\end{figure*}

Figure~\ref{fig:stage_recovery} decomposes Resume SR by the order of recovery. For repeated interruption (ABACA), it reports the first and second resumptions ($B\!\rightarrow\!A$ and $C\!\rightarrow\!A$). For nested interruption (ABCBA), it reports the inner and outer resumptions ($C\!\rightarrow\!B$ and $B\!\rightarrow\!A$). PIFR-Rank exhibits different stage-wise profiles across the three benchmarks. Under repeated interruption, its Resume SR decreases from 32.5\% to 17.5\% on PIFR-SG3D and from 53.8\% to 38.7\% on PIFR-OVON, but increases from 44.0\% to 66.0\% on PIFR-GOAT. Under nested interruption, its inner and outer Resume SRs are 1.9\% and 24.5\% on PIFR-SG3D, 50.5\% and 49.5\% on PIFR-OVON, and 56.0\% and 64.0\% on PIFR-GOAT. Thus, later or nested recovery does not produce a uniform monotonic decline. Instead, recovery performance varies with both the source benchmark and the position of the resumption within the interruption sequence.

\subsection{Ablation Study of PIFR-Rank}
\label{sec:pifr_rank_ablation}
%-----------------------------------ICLR 版本--------------------------------
%----------------------------------ICLR 版本----------------------------------
\begin{wraptable}{r}{0.65\textwidth}
\vspace{-0.5\baselineskip}
\centering
\vspace{-3mm}
\caption{\small Ablation study of PIFR-Rank. \textbf{BCE.}, \textbf{CR.} and \textbf{RW.} denote BCE Loss, candidate ranking and resumption weighting, respectively. The best result for each metric is highlighted in bold.}
\label{tab:pifr_rank_ablation}
\small
\resizebox{0.65\textwidth}{!}{
\begin{tabular}{l|cccc}
\toprule
Method
& SR$\uparrow$
& SPL$\uparrow$
& Int. SR$\uparrow$
& Res. SR$\uparrow$ \\
\midrule
Baseline
& 37.03 & 21.55 & 31.34 & 40.82 \\
\midrule
Baseline + \textbf{BCE.}
&  \textbf{39.82} & 23.32 & 33.46 & 44.91 \\
\midrule
Baseline + \textbf{BCE.} + \textbf{CR.}
&39.28 & \textbf{23.54} & 34.36 & 43.10\\
\midrule
Baseline + \textbf{BCE.} + \textbf{CR.} + \textbf{RW.}
& 39.21 & 23.25
& \textbf{35.25} & \textbf{45.04}   \\
\bottomrule
\end{tabular}}
\vspace{-0.5\baselineskip}
\end{wraptable}
%----------------------------------ICLR 版本----------------------------------
%----------------------------------Arxiv 版本----------------------------------
% \begin{table}[!htb]
% % \vspace{-0.5\baselineskip}
% \centering
% % \vspace{-3mm}
% \caption{\small Ablation study of PIFR-Rank. \textbf{BCE.}, \textbf{CR.} and \textbf{RW.} denote BCE Loss, candidate ranking and resumption weighting, respectively. The best result for each metric is highlighted in bold.}
% \label{tab:pifr_rank_ablation}
% \small
% %\setlength{\tabcolsep}{4pt}
% \resizebox{0.75\textwidth}{!}{
% \begin{tabular}{l|cccc}
% \toprule
% Method
% & SR$\uparrow$
% & SPL$\uparrow$
% & Int. SR$\uparrow$
% & Res. SR$\uparrow$ \\
% \midrule
% Baseline
% & 37.03 & 21.55 & 31.34 & 40.82 \\
% \midrule
% Baseline + \textbf{BCE.}
% &  \textbf{39.82} & 23.32 & 33.46 & 44.91 \\
% \midrule
% Baseline + \textbf{BCE.} + \textbf{CR.}
% &39.28 & \textbf{23.54} & 34.36 & 43.10\\
% \midrule
% Baseline + \textbf{BCE.} + \textbf{CR.} + \textbf{RW.}
% & 39.21 & 23.25
% & \textbf{35.25} & \textbf{45.04}   \\
% \bottomrule
% \end{tabular}}
% \vspace{-0.5\baselineskip}
% \end{table}
%----------------------------------Arxiv 版本----------------------------------
%-----------------------------------ICLR 版本--------------------------------
We conduct ablations on the Seen validation splits of PIFR-SG3D, PIFR-OVON, and PIFR-GOAT. To give each dataset and interruption structure equal weight, we report an equal-cell macro average over the nine dataset-by-structure combinations. Within each cell, SR and SPL are averaged over atomic navigation segments, whereas Interrupt SR and Resume SR are computed over segments assigned to the corresponding execution stages.

%-----------------------------------Arxiv 版本--------------------------------
% \input{Tables/ablation}
%-----------------------------------Arxiv 版本--------------------------------

As shown in Table~\ref{tab:pifr_rank_ablation}, adding candidate-level BCE supervision to the baseline improves all four metrics: SR increases from 37.03\% to 39.82\%, SPL from 21.55\% to 23.32\%, Interrupt SR from 31.34\% to 33.46\%, and Resume SR from 40.82\% to 44.91\%. Adding candidate ranking (CR) further improves SPL and Interrupt SR to 23.54\% and 34.36\%, respectively, while SR and Interrupt SR decrease slightly. Adding resumption weighting (RW) yields the highest Interrupt SR and Resume SR, at 35.25\% and 45.04\%, respectively; SR and SPL are 39.21\% and 23.25\%, below the BCE-only variant. Thus, BCE provides the broadest improvement over the baseline, while CR and RW shift the metric profile toward stronger stage-specific performance, especially during resumption.

\section{Conclusion}
We present PIFR, a benchmark for evaluating task-sequence interruption and resumption in embodied navigation. By introducing single, repeated, and nested interruptions, PIFR assesses both navigation success and recovery of unfinished tasks from the post-interruption state. Experiments show that aggregate success alone can obscure differences in interruption handling and resumption, whose performance varies across execution structures. We also introduce PIFR-Rank, a candidate-ranking baseline that improves resumption-related metrics, though gains vary across settings. PIFR provides a testbed for studying task recovery in long-horizon embodied navigation.

% We present PIFR, a benchmark for studying interruption and resumption in persistent embodied navigation. PIFR extends conventional sequential navigation by introducing various kinds of preemption. This setting allows us to evaluate not only whether an agent can navigate successfully, but also whether it can resume an unfinished task sequence from its next pending goal after intervening physical execution. Our experiments across multiple navigation settings reveal that overall navigation success does not fully characterize recovery behavior: methods with similar aggregate performance can exhibit substantially different interruption and resumption outcomes, and different execution structures expose distinct recovery profiles. Our method improves candidate selection under persistent execution and achieves gains over the baseline in most evaluated settings. We hope PIFR encourages future work toward embodied agents that can maintain and recover task intent under long-horizon, dynamically changing execution.

\bibliography{iclr2027_conference}
\bibliographystyle{iclr2027_conference}

\end{document}